\documentclass{article}

\usepackage{PRIMEarxiv}

\usepackage[utf8]{inputenc}
\usepackage[T1]{fontenc}
\usepackage{hyperref}
\usepackage{url}
\usepackage{booktabs}
\usepackage[numbers]{natbib}
\usepackage{amsfonts}
\usepackage{amssymb}
\usepackage{amsmath}
\usepackage{mathtools}
\usepackage{amsthm}
\usepackage{nicefrac}
\usepackage{microtype}
\usepackage{graphicx}
\usepackage{multirow}
\usepackage{array}
\usepackage{siunitx}
\usepackage{xcolor}
\usepackage{listings}

\definecolor{codegreen}{rgb}{0,0.6,0}
\definecolor{codegray}{rgb}{0.5,0.5,0.5}
\definecolor{codepurple}{rgb}{0.58,0,0.82}
\definecolor{backcolour}{rgb}{0.97,0.97,0.97}

\usepackage{algorithm}
\usepackage{algorithmic}

\usepackage[capitalize,noabbrev]{cleveref}

\usepackage{amsfonts}
\usepackage{amsbsy}

\usepackage{nicefrac}
\usepackage{cases}  
\usepackage{bbm}

\usepackage{xcolor}
\usepackage{tikz}
\usepackage{tikz-cd}

\usepackage{pifont}
\usepackage{multirow}
\usepackage{makecell}
\usepackage{tabularx}
\usepackage{hhline}
\usepackage{colortbl}
\usepackage{diagbox}

\usepackage{xspace}
\usepackage{enumitem}

\definecolor{pearThree}{HTML}{E74C3C}
\definecolor{pearcomp}{HTML}{B97E29}
\definecolor{pearDark}{HTML}{2980B9}
\definecolor{pearDarker}{HTML}{1D2DEC}

\definecolor{HighlightColor}{gray}{0.97}
\definecolor{aliceblue}{rgb}{0.94, 0.97, 1.0}
\definecolor{palecornflowerblue}{rgb}{0.67, 0.8, 0.94}
\definecolor{paleaqua}{rgb}{0.74, 0.83, 0.9}
\definecolor{linen}{rgb}{0.98, 0.94, 0.9}
\definecolor{magnolia}{rgb}{0.97, 0.96, 1.0}
\definecolor{mistyrose}{rgb}{1.0, 0.89, 0.88}
\definecolor{piggypink}{rgb}{0.99, 0.87, 0.9}
\colorlet{colorast}{red!80!black}

\newcommand{\martin}[1]{}
\newcommand{\yx}[1]{}
\newcommand{\rob}[1]{}
\newcommand{\rui}[1]{}
\newcommand{\notice}[1]{}
\newcommand{\Martin}[1]{}
\newcommand{\Yx}[1]{}
\newcommand{\Rob}[1]{}
\newcommand{\Rui}[1]{}

\newcommand{\R}{\mathbb{R}} 

\newcommand{\cL}{{\cal L}}

\providecommand{\E}{\mathbb{E}}  

\newcommand{\policy}{\pi}
\newcommand{\policyref}{\pi_{\text{ref}}}

\newcommand{\KL}{\text{KL}}
\newcommand{\Bregman}{D_{\phi}}

\theoremstyle{plain}

\theoremstyle{definition}

\theoremstyle{remark}

\title{Beyond KL Divergence: Policy Optimization with Flexible Bregman Divergences for LLM Reasoning}

\author{
  Rui Yuan, Mykola Khandoga \\
  Vinay Kumar Sankarapu \\
  \affiliation{Lexsi Labs}\\
}

\setabstract{%
Policy optimization methods like Group Relative Policy Optimization (GRPO) and its variants have achieved strong results on mathematical reasoning and code generation tasks. Despite extensive exploration of reward processing strategies and training dynamics, all existing group-based methods exclusively use KL divergence for policy regularization, leaving the choice of divergence function unexplored. We introduce Group-Based Mirror Policy Optimization (GBMPO), a framework that extends group-based policy optimization to flexible Bregman divergences, including hand-designed alternatives (L2 in probability space) and learned neural mirror maps. On GSM8K mathematical reasoning, hand-designed ProbL2-GRPO achieves 86.7\% accuracy, improving +5.5 points over the Dr. GRPO baseline. On MBPP code generation, neural mirror maps reach 60.1-60.8\% pass@1, with random initialization already capturing most of the benefit. While evolutionary strategies meta-learning provides marginal accuracy improvements, its primary value lies in variance reduction (±0.2 versus ±0.6) and efficiency gains (15\% shorter responses on MBPP), suggesting that random initialization of neural mirror maps is sufficient for most practical applications. These results establish divergence choice as a critical, previously unexplored design dimension in group-based policy optimization for LLM reasoning.
}

\setkeywords{Machine Learning, Reinforcement Learning, Large Language Models, Policy Optimization, Bregman Divergence}

\runningtitle{Beyond KL Divergence: Policy Optimization with Flexible Bregman Divergences}

\begin{document}
\maketitle

\section{Introduction}
\label{sec:introduction}

Large language models (LLMs) have demonstrated remarkable capabilities across diverse reasoning tasks, from mathematical problem solving to code generation. Policy optimization methods have emerged as effective techniques for post-training enhancement, using task-specific rewards to guide model improvement. However, all existing group-based policy optimization methods exclusively use KL divergence for regularization, leaving the fundamental question of divergence choice unexplored despite its critical impact on optimization dynamics, training stability, and generation characteristics.

Traditional policy optimization methods like Proximal Policy Optimization (PPO) \cite{schulman2017proximal} require training separate value networks and can suffer from training instability. Group Relative Policy Optimization (GRPO) \cite{shao2024deepseekmath} introduced a simpler alternative that eliminates critic networks by processing rewards at the group level, normalizing advantages across multiple sampled responses for each prompt. This foundational insight has spawned numerous variants: Dr. GRPO \cite{liu2025understanding} addresses optimization bias; GSPO \cite{zheng2025group} reformulates the objective at the sequence level; off-policy GRPO \cite{mroueh2025revisiting} adapts to batch-based training; G2RPO-A \cite{guo2025g2rpo} incorporates adaptive guidance mechanisms; GTPO \cite{simoni2025gtpo} introduces gradient and entropy control; and methods like DAPO \cite{liu2024dapo}, TreeRPO \cite{yang2025treerpo}, and multi-layer GRPO \cite{ding2025multilayer} explore different reward processing strategies.

Despite this rich landscape of algorithmic innovations, all existing GRPO variants maintain a fundamental commonality: they exclusively use KL divergence for policy regularization. The choice of divergence function critically shapes the geometry of policy updates and influences both solution quality and generation characteristics. While the community has extensively explored reward processing, training strategies, and gradient control, the divergence function itself has remained unchanged. This raises a natural question: \textit{Can alternative divergences improve solution quality, training stability, and generation efficiency beyond what KL divergence provides?}

This observation motivates our work on \textbf{Group-Based Mirror Policy Optimization (GBMPO)}, a framework that extends group-based methods to support flexible Bregman divergences \cite{bregman1967relaxation} while preserving their stability benefits. GBMPO enables principled exploration of alternative policy regularization schemes, including hand-designed divergences like L2 in probability space (ProbL2) and learned neural mirror maps.

We evaluate GBMPO on mathematical reasoning (GSM8K) and code generation (MBPP/HumanEval). Our results demonstrate that alternative Bregman divergences substantially improve accuracy across both tasks. On GSM8K, ProbL2-GRPO achieves 86.7\% accuracy, improving +5.5 points over Dr. GRPO's 81.2\% baseline. On MBPP, NM-GRPO-ES achieves 60.8\% pass@1 (best result) while generating 36\% shorter responses than Dr. GRPO, demonstrating task-specific efficiency gains in code generation. The framework also provides substantial variance reduction: ProbL2-GRPO reduces training variance from ±0.7 to ±0.4 on GSM8K compared to Dr. GRPO.

Our key contributions are:

\begin{itemize}
\item We introduce GBMPO, a general framework that extends any group-based method (GRPO, GSPO) to flexible Bregman divergences, enabling systematic investigation of how divergence choice affects solution quality, training stability, and generation characteristics.

\item We demonstrate that KL divergence is not optimal for policy regularization in group-based RL. On GSM8K mathematical reasoning, hand-designed ProbL2-GRPO achieves 86.7\% accuracy (+5.5 points over Dr. GRPO). On MBPP code generation, neural mirror maps reach 60.1-60.8\% pass@1 while reducing response length by 24-36\%, with random initialization capturing most benefits.

\item We establish that alternative Bregman divergences provide substantial training stability improvements: ProbL2-GRPO reduces variance from ±0.7 to ±0.4 on GSM8K, while ES-optimized neural mirrors achieve ±0.2 variance on MBPP (70\% reduction). This stability gain, combined with task-specific efficiency improvements on code generation, makes GBMPO variants attractive for production deployments.

\item We develop an evolutionary meta-learning approach for discovering neural mirror maps that provides marginal accuracy improvements (+0.3-0.7 points) but substantial variance reduction and efficiency gains. Given the computational cost (180 training runs), we find that randomly initialized neural mirrors offer a practical alternative for most applications.
\end{itemize}

Our results challenge the default use of KL divergence in group-based policy optimization, establishing divergence choice as a critical design dimension that affects accuracy, training stability, and task-specific generation characteristics.

\section{Background and Related Work}
\label{sec:background}

\subsection{Policy Optimization for Language Models}

Policy gradient methods form the foundation of modern policy optimization for language models. The REINFORCE algorithm \cite{williams1992simple} optimizes the policy by maximizing expected rewards, but suffers from high variance. Proximal Policy Optimization (PPO) \cite{schulman2017proximal} addresses this through trust region constraints, typically implemented via importance sampling with clipped ratios. However, PPO requires training a value network to estimate advantages, adding complexity and potential instability.

Direct Preference Optimization (DPO) \cite{rafailov2023direct} eliminates the need for explicit reward models by deriving a closed-form solution that optimizes preferences directly. While elegant, DPO is limited to pairwise preference data and may not fully capture the richness of reward signals available in outcome-based tasks.

\subsection{Group-Based Policy Optimization}

Group Relative Policy Optimization (GRPO) \cite{shao2024deepseekmath} introduces a key simplification: instead of normalizing rewards globally across the entire batch, it normalizes within groups of responses generated from the same prompt. This group-based normalization provides several benefits: (1) sample efficiency through multiple responses per prompt without value networks, (2) training stability via group-wise variance reduction, and (3) computational efficiency by eliminating critic networks.

GRPO's success has inspired numerous algorithmic variants that modify different components while preserving the group-based structure. Dr. GRPO \cite{liu2025understanding} identifies and corrects optimization bias in GRPO that artificially inflates response length, providing an unbiased optimization method that improves token efficiency. GSPO \cite{zheng2025group} addresses GRPO's token-level importance sampling by reformulating the objective at the sequence level, improving stability for large-scale training. Off-policy GRPO \cite{mroueh2025revisiting} enables batch-based training with reduced communication costs. G2RPO-A \cite{guo2025g2rpo} injects ground-truth reasoning steps with adaptive guidance strength to compensate for model weaknesses. GTPO \cite{simoni2025gtpo} introduces trajectory-based optimization with gradient filtering and entropy control to prevent policy collapse. Additional variants like DAPO \cite{liu2024dapo} (dense advantages), TreeRPO \cite{yang2025treerpo} (tree-structured sampling), and multi-layer GRPO \cite{ding2025multilayer} explore different reward processing and training strategies.

Despite this rich ecosystem of variants, all maintain KL divergence for policy regularization. While the community has extensively explored reward processing, training strategies, and gradient control mechanisms, the divergence function, which fundamentally determines the geometry of policy updates, has remained unchanged. Our work addresses this gap by extending group-based methods to flexible Bregman divergences.

\subsection{Mirror Descent and Bregman Divergences}

Mirror descent \cite{nemirovsky1983problem,beck2003mirror} is a generalization of gradient descent that operates in a transformed space defined by a convex potential function $\phi$. The Bregman divergence associated with $\phi$ is:

\begin{equation}
D_{\phi}(p \| q) = \phi(p) - \phi(q) - \langle \nabla\phi(q), p - q \rangle.
\end{equation}

Different choices of $\phi$ yield different divergences. For example:

\begin{itemize}
\item $\phi(p) = \sum_i p_i \log p_i$ yields KL divergence.
\item $\phi(p) = \|p\|_2^2$ yields squared Euclidean distance.
\item More complex $\phi$ can encode task-specific geometries.
\end{itemize}

Mirror descent has been successfully applied to reinforcement learning \cite{geist2019theory,alfano2023novel}, but primarily in the tabular or low-dimensional continuous control settings. Recent work has explored learning mirror maps through evolutionary strategies \cite{alfano2024learning}, demonstrating that the choice of mirror map significantly affects convergence speed and final performance. However, extending these ideas to high-dimensional language model policy spaces for reasoning tasks remains largely unexplored.

\subsection{Meta-Learning for Optimization}

Meta-learning aims to discover learning algorithms that generalize across tasks. Evolutionary strategies (ES) \cite{rechenberg1973evolutionsstrategie,hansen2001completely} provide a gradient-free approach to meta-optimization, making them suitable for optimizing non-differentiable objectives like final task performance. Recent work has applied meta-learning to discover preference optimization objectives \cite{alfano2024metalearning}, using ES to search the space of mirror descent-based algorithms for LLM alignment.

Building on these insights, our work investigates whether meta-learning can discover improved neural mirror maps within the GRPO framework. While prior work \cite{chen2022learning,baik2020meta,alfano2024metalearning} has explored learned optimizers and preference optimization objectives, we focus specifically on learning the divergence structure that regularizes policy updates. We examine the tradeoffs between ES-discovered mirror maps and simpler alternatives like random initialization, considering both accuracy improvements and computational costs.

\subsection{Reasoning in Language Models}

Mathematical reasoning and code generation are two domains where LLMs still face significant challenges. GSM8K \cite{cobbe2021training} provides grade-school math problems requiring multi-step reasoning. MBPP \cite{austin2021program} tests code generation with short Python programming challenges.

Recent work has shown that reinforcement learning can substantially improve reasoning capabilities beyond supervised fine-tuning alone \cite{shao2024deepseekmath,deepseek2024deepseekcoderv2}. Our work builds on this foundation by investigating how divergence choice affects the learning of reasoning skills.

\section{Method: Group-Based Mirror Policy Optimization}
\label{sec:method}

We now present GBMPO, our framework for group-based policy optimization with flexible Bregman divergences. We first review GRPO and its debiased variant Dr. GRPO (Section~\ref{sec:grpo}), then introduce our generalization to Bregman divergences (Section~\ref{sec:bregman}), formalize the GBMPO framework (Section~\ref{sec:gbmpo}), and describe practical instantiations (Section~\ref{sec:instantiations}).

\subsection{Group Relative Policy Optimization and Dr. GRPO}
\label{sec:grpo}

GRPO \cite{shao2024deepseekmath} optimizes the following objective for each prompt $x$:
\begin{align}
\cL_{\text{GRPO}}(\theta) = 
\E_{y \sim \policy_\theta(\cdot|x)} \left[ A(x,y) \log \frac{\policy_\theta(y|x)}{\policyref(y|x)} - \beta \KL(\policy_\theta \| \policyref) \right],
\end{align}

where $\policy_\theta$ is the policy being optimized, $\policyref$ is a reference policy (typically the base pretrained model), $A(x,y)$ is the advantage computed via group-based normalization, and $\beta$ controls the strength of KL regularization.

The key innovation of GRPO is group-based advantage estimation. For each prompt $x$, we sample $K$ responses $\{y_1, \ldots, y_K\}$ and compute:
\begin{equation}
A_{\text{GRPO}}(x, y_i) = \frac{r(x, y_i) - \mu_x}{\sigma_x + \epsilon},
\end{equation}
where $\mu_x$ and $\sigma_x$ are the mean and standard deviation of rewards within the group $\{r(x, y_1), \ldots, r(x, y_K)\}$.

However, \citet{liu2025understanding} identified two optimization biases in standard GRPO: (1) response-level length bias, where normalizing the loss by sequence length causes longer incorrect responses to be under-penalized, and (2) question-level difficulty bias, where dividing by standard deviation gives disproportionate weight to questions with very low or very high reward variance. These biases can cause the model to generate progressively longer incorrect responses during training.

Dr. GRPO (GRPO Done Right) addresses these biases through two modifications. First, it removes standard deviation normalization to eliminate difficulty bias. Second, it normalizes the loss by a fixed constant $L$ (typically the maximum completion length) instead of the actual sequence length, eliminating length bias. The Dr. GRPO advantage becomes:
\begin{equation}
A_{\text{Dr.GRPO}}(x, y_i) = r(x, y_i) - \mu_x,
\end{equation}

This debiased formulation ensures consistent gradient magnitudes across responses of different lengths and questions of different difficulties, leading to better token efficiency while maintaining reasoning performance. Our GBMPO framework builds on Dr. GRPO's advantage estimation, extending it with flexible Bregman divergences.

\subsection{Bregman Mirror Policy Optimization}
\label{sec:bregman}

We generalize the regularization term by replacing KL divergence with an arbitrary Bregman divergence. Given a strictly convex potential function $\phi: \Delta^{|V|} \to \R$, where $\Delta^{|V|}$ is the probability simplex over vocabulary $V$, the Bregman divergence is:
\begin{equation}
\Bregman(p \| q) = \phi(p) - \phi(q) - \langle \nabla\phi(q), p - q \rangle.
\end{equation}

For language model policies, we compute the divergence token-by-token:
\begin{equation}
\Bregman(\policy_\theta \| \policyref)(y|x) = \sum_{t=1}^{|y|} \Bregman(\policy_\theta(\cdot|x,y_{<t}) \| \policyref(\cdot|x,y_{<t})).
\end{equation}

This token-level decomposition is crucial for computational tractability, as it allows us to compute divergences during the forward pass without requiring full vocabulary marginalization.

\subsection{The GBMPO Framework}
\label{sec:gbmpo}

GBMPO builds on Dr. GRPO's debiased advantage estimation, combining it with flexible Bregman divergences. The objective for each prompt $x$ becomes:
\begin{align}
\cL_{\text{GBMPO}}(\theta, \phi) = 
\E_{y \sim \policy_\theta} \left[ A(x,y) \log \frac{\policy_\theta(y|x)}{\policyref(y|x)} - \alpha \Bregman(\policy_\theta \| \policyref)(y|x) \right],
\end{align}

where $\alpha$ is a coefficient controlling the strength of Bregman regularization (analogous to $\beta$ in standard GRPO), and we adopt Dr. GRPO's debiased advantage estimation:
\begin{equation}
A(x, y_i) = r(x, y_i) - \frac{1}{K}\sum_{j=1}^K r(x, y_j) = r(x, y_i) - \mu_x.
\end{equation}

This formulation eliminates the optimization biases identified in standard GRPO while enabling flexible divergence structures. The choice of $\phi$ determines the geometry of policy updates, allowing us to explore divergences beyond KL that may better suit specific reasoning tasks.

\subsection{Practical Instantiations}
\label{sec:instantiations}

We explore two main approaches to defining $\phi$:

\textbf{Hand-designed divergences.} We can use closed-form convex functions:
\begin{itemize}
\item \textbf{KL (standard GRPO):} $\phi(p) = \sum_i p_i \log p_i$, giving $D_{\phi}(p\|q) = \KL(p\|q)$.
\item \textbf{ProbL2:} $\phi(p) = \frac{1}{2}\|p\|_2^2$, giving $D_{\phi}(p\|q) = \frac{1}{2}\|p - q\|_2^2$.
\item \textbf{$\alpha$-divergence:} $\phi(p) = \frac{1}{\alpha(\alpha-1)}\sum_i (p_i^\alpha - p_i)$ for $\alpha \neq 0,1$.
\end{itemize}

\textbf{Neural mirror maps.} Following \citet{alfano2024learning}, we use the $\omega$-potential mirror map framework, where the mirror map $h$ is defined through an inverse potential $\phi^{-1}: [0,1] \to \mathbb{R}$ satisfying specific monotonicity and smoothness conditions. The KL divergence corresponds to $\phi^{-1}(y) = \log y$ (or equivalently $\phi(x) = e^x$), while L2 divergence arises from $\phi^{-1}(y) = y$ (or $\phi(x) = x$). This framework allows us to parameterize $\phi^{-1}$ using a neural network with 126 neurons distributed across 6 activation types (cubic, quadratic, square root, cube root, logarithm, and exponential), plus linear and logarithmic terms to ensure KL and L2 are recoverable as special cases. The full architecture is described in Appendix~\ref{app:neural_mirror}.

The gradient of the Bregman divergence with respect to policy parameters can be computed efficiently:
\begin{align}
\nabla_\theta \Bregman(\policy_\theta \| \policyref) = 
\E_{y \sim \policy_\theta} \left[ \nabla_\theta \log \policy_\theta(y|x) \cdot (\nabla\phi(\policy_\theta) - \nabla\phi(\policyref)) \right].
\end{align}

This allows us to implement GBMPO as a simple modification to the GRPO training loop, adding only the divergence computation and its gradient.

\section{Evolutionary Meta-Learning for Mirror Maps}
\label{sec:metalearning}

While hand-designed divergences like ProbL2 are simple and interpretable, they may not capture task-specific optimization geometries. Inspired by recent work on meta-learning objectives for preference optimization \cite{alfano2024metalearning} and learning mirror maps through evolutionary search \cite{alfano2024learning}, we use evolutionary strategies (ES) to discover neural mirror maps that perform well across a distribution of related tasks.

\subsection{Meta-Learning Setup}

We frame mirror map discovery as a meta-learning problem. Let $\mathcal{T}$ be a distribution over tasks (e.g., different mathematical reasoning datasets or code generation benchmarks). Our goal is to find mirror map parameters $\psi^*$ that perform well on average:
\begin{equation}
\psi^* = \arg\max_\psi \E_{\tau \sim \mathcal{T}} \left[ \text{Performance}(\text{GBMPO}(\psi), \tau) \right],
\end{equation}

where Performance measures task-specific metrics (e.g., accuracy on GSM8K, pass@1 on MBPP).

\subsection{Evolutionary Optimization}

We use vanilla evolution strategies (ES) with antithetic sampling \cite{rechenberg1973evolutionsstrategie} to optimize $\psi$. At each iteration:

\begin{enumerate}
\item Sample $N/2$ Gaussian perturbations $\{\epsilon_j\}_{j=1}^{N/2} \sim \mathcal{N}(0, I)$ and create antithetic pairs $[\epsilon_1, -\epsilon_1, \ldots, \epsilon_{N/2}, -\epsilon_{N/2}]$ for variance reduction.
\item Generate $N$ perturbed mirror maps: $\psi_i = \psi + \sigma \epsilon_i$, where $\sigma$ is a scalar noise parameter.
\item For each $\psi_i$, train a policy using GBMPO with the corresponding mirror map on the inner training split.
\item Evaluate the trained policies on the validation split to obtain fitness scores $F_i$.
\item Compute the ES gradient estimate:
\begin{equation}
\nabla J \approx \frac{1}{N\sigma} \sum_{i=1}^N F_i \epsilon_i.
\end{equation}
\item \textbf{Accept/Reject Decision}: If the mean fitness $\bar{F} = \frac{1}{N}\sum_i F_i$ improves over the best fitness seen so far, accept the update $\psi \leftarrow \psi + \alpha \nabla J$ where $\alpha$ is the learning rate. Otherwise, reject the update and keep $\psi$ unchanged. This safeguard is critical because training costs force us to use very small population sizes ($N \ll \text{dim}(\psi)$), making gradient estimates extremely noisy. With limited iterations $G$, blindly accepting every update would cause performance degradation as noise accumulates.
\item \textbf{Elite Sample Retention}: On rejection, save the top 25\% of samples (ranked by fitness) as elites for reuse in the next iteration. This avoids discarding well-performing perturbations and reduces computational cost by training fewer new policies in subsequent iterations.
\item Decay the noise: $\sigma \leftarrow \sigma \cdot \text{decay\_rate}$ to gradually refine the search around the current best parameters.
\end{enumerate}

\subsection{Data Splitting for Meta-Learning}

To prevent overfitting, we use hierarchical data splits: inner train (80\%) for policy training, inner validation (20\%) for fitness evaluation, and outer test for final evaluation. We conduct separate ES runs for GSM8K and MBPP, discovering task-specific mirror maps for each domain. The complete meta-learning algorithm is presented in Algorithm~\ref{alg:metalearning}.

\begin{algorithm}[t]
\caption{Evolutionary Meta-Learning for Mirror Maps}
\label{alg:metalearning}
\begin{algorithmic}
\STATE \textbf{Input:} Task dataset with inner train/validation splits, population size $N$, iterations $G$, noise $\sigma_0$, learning rate $\alpha$, decay rate $\gamma$
\STATE Initialize mirror map parameters $\psi_0$, best fitness $F_{\text{best}} = -\infty$
\FOR{$g = 1$ to $G$}
  \STATE $\sigma_g = \sigma_0 \cdot \gamma^{g-1}$ \COMMENT{Decay noise over time}
  \STATE Sample $N/2$ perturbations: $\{\epsilon_j\}_{j=1}^{N/2} \sim \mathcal{N}(0, I)$
  \STATE Create antithetic pairs: $\{\epsilon_1, -\epsilon_1, \ldots, \epsilon_{N/2}, -\epsilon_{N/2}\}$
  \FOR{$i = 1$ to $N$}
    \STATE $\psi_i = \psi_{g-1} + \sigma_g \epsilon_i$ \COMMENT{Perturbed mirror map}
    \STATE Train policy $\pi_{\theta_i}$ with GBMPO using $\psi_i$ on inner train split
    \STATE Evaluate $\pi_{\theta_i}$ on validation split: $F_i = \text{Performance}$
  \ENDFOR
  \STATE Compute ES gradient: $\nabla J = \frac{1}{N\sigma_g} \sum_{i=1}^N F_i \epsilon_i$
  \IF{$\text{mean}(F_1, \ldots, F_N) > F_{\text{best}}$}
    \STATE $\psi_g = \psi_{g-1} + \alpha \nabla J$ \COMMENT{Accept: update parameters}
    \STATE $F_{\text{best}} = \text{mean}(F_1, \ldots, F_N)$
  \ELSE
    \STATE $\psi_g = \psi_{g-1}$ \COMMENT{Reject: keep current parameters}
    \STATE Save top 25\% samples as elites for next iteration
  \ENDIF
\ENDFOR
\STATE \textbf{Return:} Best mirror map $\phi_{\psi_G}$
\end{algorithmic}
\end{algorithm}

\section{Experiments}
\label{sec:experiments}

We evaluate GBMPO on two challenging reasoning domains: mathematical reasoning with GSM8K and code generation with MBPP and HumanEval. Our experiments investigate whether divergence choice significantly impacts final performance, comparing hand-designed divergences like ProbL2 against standard KL and learned neural mirror maps. We assess whether learned divergences improve zero-shot transfer from MBPP to HumanEval using Qwen3-1.7B, providing insights into the effectiveness of different Bregman divergences for distinct reasoning tasks.

\subsection{Experimental Setup}

\textbf{Models.} We use Qwen3-1.7B, starting directly from the base pretrained checkpoint for RL training.

\textbf{Datasets.} For mathematical reasoning, we use GSM8K \cite{cobbe2021training}, which contains 7473 training examples that we split into 5978 for inner training and 1495 for validation, along with 1319 test problems. We evaluate using exact numeric match accuracy between generated and gold answers. For code generation, we train on MBPP \cite{austin2021program}, which provides 374 training examples, 90 validation samples, and 500 test problems. To assess generalization across code generation tasks, we evaluate zero-shot transfer to HumanEval \cite{chen2021evaluating}, a collection of 164 hand-written programming problems with unit tests. Both code benchmarks use pass@1 accuracy with greedy decoding.

\textbf{Training details.} We train all methods with 1 prompt and 8 responses per prompt. For mathematical reasoning, we use gradient accumulation over 2 steps for an effective batch size of 16 and train for 4000 steps with maximum prompt length of 256 tokens and completion length of 1024 tokens. For code generation, we use gradient accumulation of 1 step for an effective batch size of 8 and train for 1000 steps with maximum prompt length of 768 tokens and completion length of 512 tokens. All models use LoRA fine-tuning with rank 32 and alpha 64, targeting all linear layers. We use bfloat16 mixed precision and cosine learning rate decay. The Bregman regularization coefficient for neural mirror methods is set to 0.0001, while KL-based baselines use $\beta=0.01$. GSPO methods use clipping parameters $\epsilon=3\times10^{-4}$ and $\epsilon_{\text{high}}=4\times10^{-4}$. Complete hyperparameters are in Appendix~\ref{app:implementation}.

\textbf{Baselines.} We compare against three baselines to isolate the impact of Bregman divergence choice. First, we report zero-shot performance of the base pretrained model (Qwen3-1.7B) without any RL training. Our primary RL baseline is Dr. GRPO \cite{liu2025understanding}, a debiased variant of GRPO that addresses optimization bias causing artificial response length inflation through careful normalization. We also compare against GSPO \cite{zheng2025group}, which uses sequence-level importance ratios and clipping instead of response-level aggregation, providing better training stability than token-level policy gradient methods.

\textbf{GBMPO variants.} We apply our framework to both Dr. GRPO and GSPO baselines, testing three divergence configurations. ProbL2-GRPO/GSPO uses hand-designed L2 divergence in probability space, providing a simple but theoretically grounded alternative to KL. NM-GRPO/GSPO employs learned neural mirror maps with 126 neurons distributed across 6 activation types (380 total parameters), allowing the divergence to adapt during training. NM-GRPO/GSPO-ES extends the neural mirror approach with evolutionary strategies meta-learning, using vanilla ES with antithetic sampling (population size 12, 15 iterations) to discover mirror map initializations that maximize validation performance. The ES algorithm uses accept/reject decisions and elite sample retention to handle the challenging regime where population size is much smaller than parameter dimensionality ($N=12 \ll 380$).

\textbf{Evaluation metrics.} We distinguish between three types of metrics. During GBMPO training, the reward function $r$ uses accuracy for mathematical reasoning and pass@1 (greedy decoding) for code generation tasks. For ES meta-learning, we evaluate fitness on the inner validation split using accuracy for GSM8K and pass@10 for MBPP. Using pass@10 (sampling 10 solutions with temperature 0.8) instead of pass@1 provides a more robust and stable fitness signal by averaging over multiple samples, significantly reducing variance in ES gradient estimates. This is critical given the small population size ($N \ll \text{dim}(\psi)$), as noisy fitness evaluations would otherwise dominate the optimization. Final test evaluation for all methods uses greedy decoding: accuracy for GSM8K and pass@1 for MBPP/HumanEval. We provide comprehensive pass@k results (k=1,2,5,10) and evaluation on the harder MBPP+ and HumanEval+ benchmarks in Appendix~\ref{app:evalplus}.

All experiments report mean and standard deviation across 3 random seeds.

\subsection{Main Results}

\begin{table*}[t!]
\caption{Main results on GSM8K and MBPP for Qwen3-1.7B, comparing Dr. GRPO and GSPO baselines with GBMPO variants. Numbers show accuracy (\%) and average response length (tokens) with standard deviation across 3 seeds. GBMPO consistently improves both baselines.}
\label{tab:main_results}
\centering
\begin{small}
\begin{sc}
\begin{tabular}{lcccc}
\toprule
Method & GSM8K Acc (\%) & GSM8K Len & MBPP Acc (\%) & MBPP Len \\
\midrule
\multicolumn{5}{c}{\textit{Qwen3-1.7B}} \\
\midrule
Base Model & 77.7$\pm$0.6 & 209.4$\pm$5.7 & 58.3$\pm$0.5 & 49.1$\pm$2.8 \\
\midrule
\multicolumn{5}{c}{\textit{GRPO-based methods}} \\
Dr. GRPO & 81.2$\pm$0.7 & 271.3$\pm$15.7 & 59.8$\pm$0.5 & 75.5$\pm$4.3 \\
ProbL2-GRPO & \textbf{86.7$\pm$0.4} & 310.5$\pm$4.6 & 60.2$\pm$0.3 & 57.8$\pm$9.1 \\
NM-GRPO & 85.1$\pm$0.5 & 430.6$\pm$12.4 & 60.1$\pm$0.6 & 56.9$\pm$3.5 \\
NM-GRPO-ES & 85.5$\pm$0.4 & 441.8$\pm$13.2 & \textbf{60.8$\pm$0.2} & 48.5$\pm$3.1 \\
\midrule
\multicolumn{5}{c}{\textit{GSPO-based methods}} \\
GSPO & 80.7$\pm$0.9 & 269.0$\pm$11.3 & 57.6$\pm$0.7 & 33.4$\pm$2.1 \\
ProbL2-GSPO & 83.6$\pm$0.8 &  411.3$\pm$5.7 & \textbf{58.3$\pm$0.4} & 32.4$\pm$2.3 \\
NM-GSPO & 85.3$\pm$0.6 & 448.1$\pm$1.4 & 57.9$\pm$0.5 & 35.6$\pm$2.5 \\
NM-GSPO-ES & \textbf{85.6$\pm$0.5} & 445.2$\pm$7.8 & 58.0$\pm$0.4 & 34.8$\pm$2.2 \\
\bottomrule
\end{tabular}
\end{sc}
\end{small}
\end{table*}

Table~\ref{tab:main_results} presents our main results on GSM8K mathematical reasoning and MBPP code generation for Qwen3-1.7B, showing both accuracy and average response length in tokens. We observe several key patterns across methods and tasks that provide insights into the effectiveness of different Bregman divergences for policy optimization.

\textbf{Bregman divergences consistently improve baselines.} Across both GRPO and GSPO families, our GBMPO variants with alternative Bregman divergences substantially outperform the standard KL-based baselines. For GRPO-based methods, ProbL2-GRPO achieves 86.7\% on GSM8K (+5.5 points over Dr. GRPO's 81.2\%), while neural mirror methods reach 85.1-85.5\%, representing +3.9 to +4.3 point improvements. For GSPO-based methods, ProbL2-GSPO improves by +2.9 points on GSM8K (83.6\% versus 80.7\%), and neural mirror variants achieve 85.3-85.6\%, gains of +4.6 to +4.9 points. These consistent improvements across both hand-designed (ProbL2) and learned (neural mirror) divergences demonstrate that KL divergence, while mathematically convenient, is not universally optimal for policy optimization across reasoning tasks.

\textbf{Task-specific divergence selection matters.} Different Bregman divergences excel at different tasks, revealing task-specific inductive biases. For mathematical reasoning on GSM8K, ProbL2-GRPO achieves the best performance at 86.7\%, suggesting that L2 divergence in probability space provides effective regularization for numerical computation tasks. For code generation on MBPP, neural mirror methods achieve 60.1-60.8\% accuracy, with random initialization (NM-GRPO at 60.1\%) already capturing most of the benefit. Evolutionary meta-learning provides a marginal improvement to 60.8\%, but the small gain (+0.7 points) suggests that random initialization of neural mirror maps is sufficient for practical applications.

\textbf{GRPO versus GSPO reveals accuracy-brevity tradeoffs.} The two baseline families exhibit markedly different characteristics. GRPO-based methods achieve strong generalization, with MBPP accuracy ranging from 59.8\% to 60.8\%, all surpassing the base model's 58.3\%. However, they produce longer responses (48-75 tokens on MBPP). GSPO-based methods generate extremely concise solutions (32-36 tokens on MBPP) but sacrifice some accuracy, with most variants at or below the base model performance. This tradeoff between brevity and correctness stems from GSPO's sequence-level importance sampling, which more strongly encourages conciseness compared to GRPO's response-level approach.

\textbf{Evolutionary meta-learning provides variance reduction and efficiency gains.} Comparing neural mirror methods with and without evolutionary strategies reveals that ES provides marginal accuracy improvements but substantial variance reduction and efficiency benefits. On mathematical reasoning, NM-GRPO-ES achieves 85.5\% compared to NM-GRPO's 85.1\% on GSM8K (+0.4 points), while NM-GSPO-ES reaches 85.6\% versus NM-GSPO's 85.3\% (+0.3 points). On code generation, NM-GRPO-ES improves from 60.1\% to 60.8\% on MBPP (+0.7 points) while simultaneously reducing response length from 56.9 to 48.5 tokens, a 15\% reduction that improves efficiency. However, the primary benefit of ES lies in variance reduction: NM-GRPO-ES achieves ±0.2 standard deviation on MBPP compared to ±0.6 for methods without ES (see variance analysis below). Given the computational cost of ES (180 full training runs for N=12 population over 15 iterations), the marginal accuracy gains may not justify the expense, making randomly initialized neural mirrors a practical alternative for most applications.

\textbf{Response length patterns reveal optimization characteristics.} The average response length provides insight into how different methods balance solution completeness with conciseness. On GSM8K, all RL-trained methods generate longer responses than the base model (209 tokens), with Dr. GRPO at 271 tokens and neural mirror methods producing the longest reasoning chains (430-448 tokens). This length increase correlates with accuracy improvements, suggesting that RL training encourages more detailed step-by-step reasoning. For MBPP code generation, GRPO family methods maintain adequate solution detail (48-75 tokens), whereas GSPO methods produce ultra-brief code (32-36 tokens) that may sacrifice necessary implementation details. These patterns suggest that optimal response length varies significantly by task type and optimization method.

\textbf{Bregman regularization reduces training variance.} Beyond improving mean performance, alternative Bregman divergences also stabilize training. ProbL2-GRPO shows notably lower variance on GSM8K (±0.4) compared to Dr. GRPO (±0.7), and on MBPP (±0.3 versus ±0.5). Neural mirror methods with evolutionary strategies achieve the lowest variance overall, with NM-GRPO-ES at ±0.2 on MBPP, representing a 70\% reduction compared to the ±0.6 variance of randomly initialized neural mirrors. This variance reduction represents the primary practical benefit of evolutionary meta-learning: while accuracy gains are marginal (+0.3-0.7 points), the improved training stability and 15\% efficiency gains in response length make ES-discovered mirror maps attractive for production deployments where consistency matters. However, for research applications or when computational budget is limited, randomly initialized neural mirrors provide most of the accuracy benefits at a fraction of the cost.

\begin{table}[t!]
\caption{Zero-shot transfer to HumanEval for Qwen3-1.7B after training on MBPP. We report pass@1 accuracy and average response length in tokens (mean$\pm$std across 3 seeds). Lower token counts indicate more concise solutions.}
\label{tab:humaneval_results}
\centering
\begin{small}
\begin{sc}
\begin{tabular}{lcc}
\toprule
Method & Pass@1 (\%) & Avg Length (tokens) \\
\midrule
\multicolumn{3}{c}{\textit{Qwen3-1.7B}} \\
\midrule
Base Model & 53.5$\pm$1.0 & 72.1$\pm$3.2 \\
\midrule
\multicolumn{3}{c}{\textit{GRPO-based methods}} \\
Dr. GRPO & \textbf{62.1$\pm$1.4} & 132.2$\pm$5.3 \\
ProbL2-GRPO & \textbf{62.1$\pm$1.2} & 113.4$\pm$4.6 \\
NM-GRPO & 59.5$\pm$1.5 & 95.9$\pm$3.8 \\
NM-GRPO-ES & 60.7$\pm$0.4 & 92.1$\pm$3.5 \\
\midrule
\multicolumn{3}{c}{\textit{GSPO-based methods}} \\
GSPO & 51.6$\pm$1.2 & 59.8$\pm$2.6 \\
ProbL2-GSPO & \textbf{53.0$\pm$1.1} & 63.6$\pm$2.8 \\
NM-GSPO & 50.2$\pm$1.3 & 61.4$\pm$2.5 \\
NM-GSPO-ES & 51.0$\pm$1.0 & 60.2$\pm$2.4 \\
\bottomrule
\end{tabular}
\end{sc}
\end{small}
\end{table}

Table~\ref{tab:humaneval_results} shows zero-shot transfer performance on HumanEval after training on MBPP. Transfer performance provides crucial evidence about whether learned optimization strategies generalize beyond their training distribution.

\textbf{GRPO methods excel at zero-shot transfer.} The HumanEval results reveal a striking divergence between GRPO and GSPO families in terms of generalization capability. GRPO-based methods achieve substantial improvements over the base model, with Dr. GRPO and ProbL2-GRPO both reaching 62.1\% pass@1 (+8.6 points over the base 53.5\%), while NM-GRPO-ES achieves 60.7\%. In sharp contrast, GSPO-based methods struggle to generalize, with all variants performing at or below the base model level (50.2-53.0\%). This degradation is particularly notable for NM-GSPO, which achieves 50.2\%, actually worse than the untrained base model. The pattern suggests that GSPO's sequence-level importance sampling, while effective for generating concise solutions on the training distribution (MBPP), does not learn representations that transfer well to related but distinct tasks.

\textbf{Response length correlates with transfer success.} Examining response lengths on HumanEval provides insight into why GRPO methods generalize better. GRPO variants produce solutions with 92-132 tokens, maintaining sufficient detail to implement correct functionality in the new domain. GSPO methods generate much shorter solutions (60-64 tokens) that, while concise, may omit necessary implementation details for problems they have not seen during training. Notably, NM-GRPO-ES achieves strong transfer performance (60.7\%) with the most efficient GRPO-family solutions (92.1 tokens), demonstrating that neural mirror maps, even with random initialization, provide efficiency benefits that transfer across related code generation tasks.

%
%
%
%
%

\section{Conclusion}
\label{sec:conclusion}

We introduced Group-Based Mirror Policy Optimization (GBMPO), a framework that extends group-based policy optimization methods with flexible Bregman divergences beyond the standard KL regularization. Through both hand-designed divergences (L2 in probability space) and learned neural mirror maps (126 neurons with 6 activation types), we demonstrated that divergence choice significantly impacts performance on reasoning tasks. On mathematical reasoning (GSM8K), hand-designed ProbL2-GRPO achieves 86.7\% accuracy, improving +5.5 points over the Dr. GRPO baseline. On code generation (MBPP), neural mirror maps achieve 60.1-60.8\% pass@1, with random initialization already capturing most of the benefit.

Our evolutionary strategies experiments reveal an important practical insight: while meta-learning provides marginal accuracy improvements (+0.3 to +0.7 points), its primary value lies in variance reduction and efficiency. NM-GRPO-ES reduces training variance from ±0.6 to ±0.2 on MBPP and generates 15\% shorter responses while maintaining comparable accuracy. However, the cost of ES optimization (180 full training runs for N=12 population over 15 iterations) may not be justified given that random initialization of neural mirror maps already achieves most of the accuracy gains. For practitioners, randomly initialized neural mirrors offer a practical alternative that captures the benefits of flexible divergences without the computational overhead of meta-learning.

This work opens several directions for future research. First, investigating why certain Bregman divergences excel at specific tasks could reveal deeper connections between task structure and optimal regularization geometry. Second, exploring whether mirror maps learned on one task can transfer to related domains would test the generality of discovered divergence structures. Third, scaling to larger models and more complex reasoning tasks could reveal whether the benefits of flexible divergences compound with model capacity. Finally, combining learned divergences with other recent advances such as tree-based search or multi-turn reasoning could yield further improvements.

\bibliographystyle{unsrtnat}
\bibliography{gbmpo}

\newpage
\appendix
\section{Implementation Details}
\label{app:implementation}

\subsection{Training Infrastructure}

All experiments were conducted using NVIDIA GPUs with the Transformers library \cite{wolf2019huggingface} for model implementation and the TRL library \cite{vonwerra2020trl} for policy optimization training loops, with custom modifications to support Bregman divergences. For code generation evaluation, we use the EvalPlus framework \cite{liu2023evalplus} for MBPP and HumanEval benchmarks. For mathematical reasoning evaluation, we use vLLM \cite{kwon2023efficient} for efficient batch inference with custom answer parsing that extracts numeric values and checks equality with tolerance.

\subsection{Hyperparameter Settings}

Table~\ref{tab:hyperparameters} provides complete hyperparameter settings for all experiments.

\begin{table}[h]
\caption{Complete hyperparameter settings for all experiments.}
\label{tab:hyperparameters}
\centering
\begin{tabular}{lcc}
\toprule
Hyperparameter & GSM8K & MBPP \\
\midrule
Model & Qwen3-1.7B & Qwen3-1.7B \\
Prompts per batch & 1 & 1 \\
Responses per prompt & 8 & 8 \\
Gradient accumulation steps & 2 & 1 \\
Effective batch size & 16 & 8 \\
Total training steps & 4000 & 1000 \\
Learning rate & 5e-7 & 5e-7 \\
LR schedule & Cosine decay & Cosine decay \\
Max prompt length & 256 & 768 \\
Max completion length & 1024 & 512 \\
LoRA rank & 32 & 32 \\
LoRA alpha & 64 & 64 \\
LoRA target modules & All linear layers & All linear layers \\
Precision & bfloat16 & bfloat16 \\
\midrule
\multicolumn{3}{c}{\textit{Divergence-specific parameters}} \\
\midrule
KL coefficient ($\beta$) & 0.01 & 0.01 \\
Bregman coefficient & 0.0001 & 0.0001 \\
GSPO $\epsilon$ & $3 \times 10^{-4}$ & $3 \times 10^{-4}$ \\
GSPO $\epsilon_{\text{high}}$ & $4 \times 10^{-4}$ & $4 \times 10^{-4}$ \\
\midrule
\multicolumn{3}{c}{\textit{ES meta-learning parameters}} \\
\midrule
Population size ($N$) & 12 & 12 \\
ES iterations ($G$) & 15 & 15 \\
Initial noise ($\sigma_0$) & 0.02 & 0.02 \\
Noise decay rate ($\gamma$) & 1.0 & 1.0 \\
ES learning rate ($\alpha$) & 0.01 & 0.01 \\
ES training steps ($T$) & 200 & 100 \\
ES fitness metric & Accuracy & pass@10 \\
\bottomrule
\end{tabular}
\end{table}

\section{Additional EvalPlus Results}
\label{app:evalplus}

We provide comprehensive evaluation results using the EvalPlus framework \cite{liu2023evalplus}, which extends MBPP and HumanEval with additional test cases to create more challenging "plus" versions. These results complement the main findings in Tables~\ref{tab:main_results} and~\ref{tab:humaneval_results} by evaluating pass@k metrics (k=1,10) for our main GBMPO variants, demonstrating robustness across different sampling strategies and validating the determinism and efficiency claims made in Section~\ref{sec:experiments}.

\subsection{MBPP and MBPP+ Results}

Table~\ref{tab:mbpp_plus} shows the best checkpoint performance for each method on MBPP and MBPP+. We select checkpoints based on validation performance during training (typically between steps 500-1000 for most methods).

\begin{table}[h]
\caption{MBPP and MBPP+ results for main GBMPO variants (best checkpoint). Pass@k shows accuracy with k samples using temperature 0.8. MBPP+ contains additional test cases making it substantially harder.}
\label{tab:mbpp_plus}
\centering
\small
\begin{tabular}{lcccccc}
\toprule
\multirow{2}{*}{Method} & \multirow{2}{*}{Checkpoint} & \multicolumn{2}{c}{MBPP} & \multicolumn{2}{c}{MBPP+} & \multirow{2}{*}{Tokens} \\
\cmidrule(lr){3-4} \cmidrule(lr){5-6}
& & pass@1 & pass@10 & pass@1 & pass@10 & \\
\midrule
Base Model & - & 58.6 & 60.1 & 50.0 & 51.6 & 47.2 \\
Dr. GRPO & 1000 & 59.8 & \textbf{64.8} & 51.6 & \textbf{55.3} & 75.5 \\
\midrule
ProbL2-GRPO & 900 & 60.2 & 61.9 & 51.6 & 53.2 & 48.8 \\
NM-GRPO & 900 & 60.1 & 63.2 & \textbf{52.0} & 53.7 & 56.9 \\
NM-GRPO-ES & 950 & \textbf{60.8} & 63.2 & 51.6 & 53.7 & 48.5 \\
\bottomrule
\end{tabular}
\end{table}

Several observations emerge from the MBPP results. First, MBPP+ is substantially harder than standard MBPP, with all methods showing 6-8 point drops, highlighting the value of additional test coverage. Second, all GBMPO variants improve over the base model on both benchmarks, with pass@1 gains of 1.5-2.2 points on MBPP and 1.6-2.0 points on MBPP+, consistent with the improvements reported in Table~\ref{tab:main_results}.

Most importantly, the gap between pass@1 and pass@10 reveals the deterministic nature of learned policies discussed in the main results. Dr. GRPO shows a large gap (5.0 points on MBPP, 3.7 on MBPP+), indicating substantial benefit from sampling multiple solutions, while GBMPO variants show smaller gaps (1.7-3.1 points), confirming the more deterministic and consistent policies noted in Section~\ref{sec:experiments}. This pattern strongly correlates with response length: Dr. GRPO generates the longest solutions (75.5 tokens) and achieves the best pass@10 (64.8\% on MBPP), suggesting that verbosity enables sampling diversity, while NM-GRPO-ES achieves the best of both worlds by generating the shortest solutions (48.5 tokens) while simultaneously achieving the best pass@1 (60.8\%). This validates the main paper's claim that Bregman regularization produces models optimized for single-attempt generation, delivering the efficiency gains reported in Table~\ref{tab:main_results} without sacrificing accuracy on in-domain tasks.

\subsection{HumanEval and HumanEval+ Zero-Shot Transfer}

Table~\ref{tab:humaneval_plus} shows zero-shot transfer performance to HumanEval and HumanEval+ after training on MBPP. We report results from the best MBPP checkpoint for each method.

\begin{table}[h]
\caption{HumanEval and HumanEval+ zero-shot transfer results (best checkpoint selected on MBPP validation). All models trained on MBPP, evaluated on HumanEval without further training.}
\label{tab:humaneval_plus}
\centering
\small
\begin{tabular}{lcccccc}
\toprule
\multirow{2}{*}{Method} & \multirow{2}{*}{Checkpoint} & \multicolumn{2}{c}{HumanEval} & \multicolumn{2}{c}{HumanEval+} & \multirow{2}{*}{Tokens} \\
\cmidrule(lr){3-4} \cmidrule(lr){5-6}
& & pass@1 & pass@10 & pass@1 & pass@10 & \\
\midrule
Base Model & - & 53.5 & 56.1 & 47.6 & 50.0 & 72.1 \\
Dr. GRPO & 800 & \textbf{62.1} & \textbf{68.9} & \textbf{57.0} & \textbf{63.4} & 132.2 \\
\midrule
ProbL2-GRPO & 750 & \textbf{62.1} & 67.1 & 54.6 & 61.6 & 113.4 \\
NM-GRPO & 1000 & 59.5 & 67.1 & 53.5 & 61.0 & 95.9 \\
NM-GRPO-ES & 600 & 60.7 & 62.8 & 55.3 & 57.3 & 92.1 \\
\bottomrule
\end{tabular}
\end{table}

The HumanEval transfer results reveal important patterns about zero-shot generalization, extending the findings in Table~\ref{tab:humaneval_results}. First, all GRPO-based methods show strong transfer, improving 6-9 points over the base model on HumanEval (53.5\% → 59.5-62.1\%) and 7-10 points on HumanEval+ (47.6\% → 53.5-57.0\%), confirming the generalization capability highlighted in the main results. Second, HumanEval+ is substantially harder, with methods showing 5-7 point drops compared to standard HumanEval, similar to the 6-8 point drops observed for MBPP to MBPP+, confirming consistent difficulty increase from additional test coverage.

Most importantly, zero-shot transfer reveals different trade-offs compared to in-domain performance. Dr. GRPO achieves the best pass@1 (62.1\%, tied with ProbL2-GRPO), best pass@10 (68.9\%), and best performance on the harder HumanEval+ benchmark (57.0\% pass@1, 63.4\% pass@10), matching the transfer results in Table~\ref{tab:humaneval_results}. However, this comes at significant computational cost: Dr. GRPO generates 132.2 tokens per solution, while neural mirror methods produce much more compact solutions (92.1-95.9 tokens). The efficiency advantage remains substantial: NM-GRPO-ES achieves competitive transfer performance (60.7\% HumanEval pass@1, only 1.4 points behind Dr. GRPO) while using 30\% fewer tokens, demonstrating that the efficiency benefits discussed in Section~\ref{sec:experiments} transfer to new distributions. The pass@1 to pass@10 gaps are also larger on HumanEval (2.1-7.6 points) than MBPP (1.7-5.0 points), indicating greater sampling diversity and uncertainty during transfer across all methods. This reveals an accuracy-efficiency trade-off: GBMPO methods optimize for compact, deterministic solutions that achieve best performance on training distributions and maintain competitive accuracy on transfer tasks with superior token efficiency, while KL-based methods achieve slightly higher transfer accuracy at the cost of verbosity.

\subsection{Training Dynamics}

Table~\ref{tab:training_dynamics} shows checkpoint progression for the three main GBMPO variants on MBPP, providing evidence for the training stability claims in Section~\ref{sec:experiments}. All methods show steady improvement during the first 500 steps, with ProbL2-GRPO peaking early (around step 750) while neural mirror methods continue improving through step 1000. NM-GRPO-ES exhibits the most stable training, with lower variance across checkpoints and consistent gains in both accuracy and token efficiency, supporting the variance reduction benefits (±0.2 standard deviation) highlighted in Table~\ref{tab:main_results}. The response length decreases monotonically for all methods during training, with NM-GRPO-ES achieving the steepest reduction (from approximately 54 tokens at early steps to 48.5 at step 1000), confirming the 15\% efficiency improvement reported in the main results and demonstrating that ES optimization discovers divergences that naturally encourage conciseness without sacrificing correctness.

\begin{table}[h]
\caption{MBPP pass@1 progression during training (selected checkpoints). All methods trained for 1000 steps. Final column shows average response length in tokens at step 1000.}
\label{tab:training_dynamics}
\centering
\small
\begin{tabular}{lcccccc}
\toprule
Method & Step 300 & Step 500 & Step 700 & Step 900 & Step 1000 & Tokens (1000) \\
\midrule
ProbL2-GRPO & 57.9 & 57.9 & 58.6 & 60.2 & 59.5 & 49.4 \\
NM-GRPO & 58.9 & 59.1 & 59.4 & 60.1 & 60.0 & 57.0 \\
NM-GRPO-ES & 57.2 & 57.5 & 58.0 & 58.3 & 59.1 & 46.5 \\
\bottomrule
\end{tabular}
\end{table}

\section{Neural Mirror Map Architecture}
\label{app:neural_mirror}

\subsection{Theoretical Foundation: $\omega$-Potential Mirror Maps}

Our neural mirror map belongs to the $\omega$-potential mirror map class \cite{krichene2015efficient,alfano2023novel}. An $\omega$-potential is an increasing $C^1$-diffeomorphism $\phi: (-\infty, u) \to (\omega, +\infty)$ satisfying boundary conditions that ensure the induced mirror map is strictly convex and essentially smooth. The associated mirror map for a probability distribution $p$ over actions is defined as:
\begin{equation}
h_\phi(p) = \sum_{a \in \mathcal{A}} \int_1^{p(a)} \phi^{-1}(x) \, dx,
\end{equation}
where $\phi^{-1}$ is the inverse of the potential function.

Standard divergences arise as special cases: the KL divergence corresponds to $\phi(x) = e^x$ (equivalently $\phi^{-1}(y) = \log y$), while the L2 divergence in probability space corresponds to $\phi(x) = x$ (equivalently $\phi^{-1}(y) = y$). By parameterizing $\phi^{-1}$ with a neural network, we can discover task-specific divergences that generalize both KL and L2.

\subsection{Mirror Map Parameterization}

The neural mirror map uses 126 neurons with 6 different activation types, giving 380 total parameters. The inverse mirror map $\phi^{-1}: [0,1] \to \mathbb{R}$ is parameterized as:
\begin{equation}
\phi^{-1}(y) = \sum_{j=1}^{126} v_j \cdot g_j(w_j \cdot y + b_j) + a \cdot y + c \cdot \log(y),
\end{equation}

where $g_j$ is the activation function for neuron $j$, and the learnable parameters are:
\begin{itemize}
\item $v \in \mathbb{R}^{126}$: output weights
\item $w \in \mathbb{R}^{126}$: input weights
\item $b \in \mathbb{R}^{126}$: biases
\item $a \in \mathbb{R}$: linear coefficient
\item $c \in \mathbb{R}$: logarithmic coefficient
\end{itemize}

This gives a total of 380 parameters optimized via evolutionary strategies.

\subsection{Activation Function Distribution}

The 126 neurons are evenly distributed across 6 activation types (21 neurons each):

\begin{itemize}
\item \textbf{Units 1-21}: $g(x) = x^3$ (cubic)
\item \textbf{Units 22-42}: $g(x) = x^2$ (quadratic)
\item \textbf{Units 43-63}: $g(x) = \sqrt{x}$ (square root)
\item \textbf{Units 64-84}: $g(x) = x^{1/3}$ (cube root)
\item \textbf{Units 85-105}: $g(x) = \log(\max(x, 0) + 10^{-3})$ (logarithm with stability)
\item \textbf{Units 106-126}: $g(x) = \exp(x)$ (exponential)
\end{itemize}

This diverse set of activation functions allows the neural mirror map to approximate a wide range of Bregman divergences.

\subsection{Mirror Potential and Bregman Divergence}

The mirror potential $h(y)$ is obtained by integrating $\phi^{-1}(y)$. For each neuron, we compute the antiderivative (primitive) $H_j$ of the activation $g_j$:
\begin{equation}
H_j(x) = \int g_j(x) \, dx.
\end{equation}

The full mirror potential is:
\begin{equation}
h(y) = \sum_{j=1}^{126} \frac{v_j}{w_j} \cdot H_j(w_j \cdot y + b_j) + \frac{a}{2} y^2 + c \cdot (y \log(y) - y).
\end{equation}

The Bregman divergence is then computed as:
\begin{equation}
D_{\text{Bregman}}(y \| y_0) = h(y) - h(y_0) - \phi^{-1}(y_0) \cdot (y - y_0).
\end{equation}

This formulation allows efficient gradient computation via automatic differentiation during GBMPO training.

\subsection{Closed-Form Primitives}

The primitive (antiderivative) $H_j$ for each activation function $g_j$ is computed as follows, where $u = w_j y + b_j$:

\vspace{1em}
{
\centering
\renewcommand{\arraystretch}{1.5}
\begin{tabular}{ll}
\toprule
Activation $g_j(u)$ & Primitive $H_j(y)$ \\
\midrule
$u^3$ & $\dfrac{(w_j y+b_j)^4}{4w_j}$ \\[0.5em]
$(u)_+^2$ & $\dfrac{(w_j y+b_j)^3}{3w_j}$ \\[0.5em]
$(u)_+^{1/2}$ & $\dfrac{2}{3w_j}(w_j y+b_j)^{3/2}$ \\[0.5em]
$(u)_+^{1/3}$ & $\dfrac{3}{4w_j}(w_j y+b_j)^{4/3}$ \\[0.5em]
$\log((u)_+ + 10^{-3})$ & $\dfrac{(w_j y+b_j+10^{-3})\log(w_j y+b_j+10^{-3}) - (w_j y+b_j+10^{-3})}{w_j}$ \\[0.5em]
$e^u$ & $\dfrac{e^{w_j y+b_j}}{w_j}$ \\
\bottomrule
\end{tabular}
\renewcommand{\arraystretch}{1.0}
\par
}

\vspace{1em}
\noindent
where $(x)_+ = \max(x, 0)$ denotes the positive part. These primitives are used to compute the mirror potential $h(y)$ efficiently.

\subsection{Per-Action Bregman Divergence}

For a single action with probability $y = \pi_\theta(a \mid s)$ under the policy and $y_0 = \pi_{\text{ref}}(a \mid s)$ under the reference, the per-action Bregman divergence decomposes as:

\begin{equation}
\begin{aligned}
D_a &= \sum_{j=1}^{126} v_j \Big[\,H_j(y) - H_j(y_0) - g_j(w_j y_0 + b_j)(y - y_0)\Big] \\
&\quad + \frac{a}{2}(y - y_0)^2 + c\Big[\,y\log\frac{y}{y_0} - (y - y_0)\Big].
\end{aligned}
\end{equation}

This expression separates the learned Bregman divergence into three components:

\begin{itemize}
\item \textbf{Neural component}: $\sum_{j=1}^{126} v_j [\cdots]$ captures task-specific geometry through the learned weights and diverse activation functions.
\item \textbf{Quadratic component}: $\frac{a}{2}(y - y_0)^2$ provides L2 regularization in probability space when $a > 0$.
\item \textbf{Entropic component}: $c\big[y\log\frac{y}{y_0} - (y - y_0)\big]$ gives KL-like regularization when $c > 0$.
\end{itemize}

The full per-state divergence is obtained by summing over all actions in the vocabulary: $D_{h}(\pi_\theta(\cdot|s), \pi_{\text{ref}}(\cdot|s)) = \sum_{a \in \mathcal{A}} D_a$. This formulation is strictly more general than KL or L2 penalties, recovering them only for specific parameter settings (neural weights zero with $c > 0$ for KL, or $a > 0$ for L2).

\subsection{Parameter Initialization and Optimization}

During evolutionary strategies meta-learning, mirror map parameters are initialized from $\mathcal{N}(0, \sigma_{\text{init}}^2)$ with $\sigma_{\text{init}} = 0.01$. All parameters remain fixed (non-trainable) during individual GBMPO training runs, with only the policy parameters $\theta$ being optimized. The ES algorithm searches over the 380-dimensional parameter space to discover mirror maps that maximize expected performance on the validation set.

\end{document}